# Cooperation Learning Enhanced Colonic Polyp Segmentation Based on Transformer-CNN Fusion

Yuanyuan Wang, Zhaohong Deng, Senior Member, IEEE, Qiongdan Lou, Shudong Hu, Kup-sze Choi, Shitong Wang

***Abstract*— Traditional segmentation methods for colonic polyps are mainly designed based on low-level features. They could not accurately extract the location of small colonic polyps. Although the existing deep learning methods can improve the segmentation accuracy, their effects are still unsatisfied. To meet the above challenges, we propose a hybrid network called Fusion-Transformer-HardNetMSEG (i.e., Fu-TransHNet) in this study. Fu-TransHNet uses deep learning of different mechanisms to fuse each other and is enhanced with multi-view collaborative learning techniques. Firstly, the Fu-TransHNet utilizes the Transformer branch and the CNN branch to realize the global feature learning and local feature learning, respectively. Secondly, a fusion module is designed to integrate the features from two branches. The fusion module consists of two parts: 1) the Global-Local Feature Fusion (GLFF) part and 2) the Dense Fusion of Multi-scale features (DFM) part. The former is built to compensate the feature information mission from two branches at the same scale; the latter is constructed to enhance the feature representation. Thirdly, the above two branches and fusion modules utilize multi-view cooperative learning techniques to obtain their respective weights that denote their importance and then make a final decision comprehensively. Experimental results showed that the Fu-TransHNet network was superior to the existing methods on five widely used benchmark datasets. In particular, on the ETIS-LaribPolypDB dataset containing many small-target colonic polyps, the mDice obtained by Fu-TransHNet were 12.4% and 6.2% higher than the state-of-the-art methods HardNet-MSEG and TransFuse-s, respectively.**

This work was supported in part by the National key R & D plan under Grant (2022YFE0112400), the NSFC under Grant 62176105, the Six Talent Peaks Project in Jiangsu Province under Grant XYDXX-056, the Hong Kong Research Grants Council (PolyU 152006/19E), the Project of Strategic Importance of the Hong Kong Polytechnic University (1-ZE1V) and the Postgraduate Research & Practice innovation Program of Jiangsu Province under Grant KYCX22_2313. (Corresponding author: Zhaohong Deng).

Y. Wang, Q. Lou and S. T. Wang are with the School of Artificial Intelligence and Computer Science, Jiangnan University and Jiangsu Key Laboratory of Media Design and Software Technology, Wuxi 214122, China. (e-mail:6201613009@stu.jiangnan.edu.cn; 6171610005@stu.jiangnan.edu.cn; wxwangst@aliyun.com).

Z. Deng is with the School of Artificial Intelligence and Computer Science, Jiangnan University, Wuxi 214122, China, and Key Laboratory of Computational Neuroscience and Brain-Inspired Intelligence (LCNBI) and ZJLab, Shanghai 200433, China. (e-mail: dengzhaohong@jiangnan.edu.cn).

S Hu is with Department of Radiology, Affiliated Hospital of Jiangnan University, 1000 Hefeng Road, Wuxi, 214122, People's Republic of China (e-mail:hsd2001054@163.com).

Q. Lou and K. S. Choi are with the School of the Centre for Smart Health, the Hong Kong Polytechnic University, Hong Kong. (e-mail: 6171610005@stu.jiangnan.edu.cn; thomasks.choi@polyu.edu.hk).

## I INTRODUCTION

Colorectal cancer (CRC) is one of the three most common cancers in the world [1]. The key to effective prevention and diagnosis is colonoscopy [2]. Segmenting colonic polyps from colonoscopy images and diagnosing accurately whether they are diseased is an import work for early treatment of CRC. In general, the color of colonic polyps is similar to that of surrounding areas, and there is interference from intestinal mucus. Therefore, the boundary of colonic polyps is often fuzzy. In this case, it is difficult to achieve accurate segmentation for colonic polyps, especially small-target colonic polyps [3]. Therefore, it is a challenging and meaningful work to improve the segmentation accuracy of colonic polyps.

There are two main types of colonic polyp segmentation methods. One is the traditional segmentation method based on low-level features such as image shape and texture [4], [5]. This kind of methods relies on medical knowledge. The other is the segmentation method based on deep learning. This kind of methods automatically captures image features.

For the traditional segmentation method, J. Bernal et al. [6] designed an automatic segmentation method according to the elliptic appearance of colonic polyps. D. C. Cheng et al. [7] distinguished colonic polyps and backgrounds according to color and texture features. However, due to the low color contrast between the two, the segmentation accuracy of this method is limited.

For the deep learning segmentation method, Q. Li et al. [8] proposed an improved fully convolutional network for colonic polyp segmentation. In order to solve the problem of boundary constraints and multi-scale feature aggregation, Y. Fang et al. [9] proposed a convolutional neural network with selective feature fusion to achieve colonic polyp segmentation. Considering the different shapes and sizes of colonic polyps, as well as the low contrast between polyps and background, R. Zhang et al. [10] constructed a convolutional neural network to adaptively processes different context information with the help of attention mechanism, global context module, and adaptive selection module. B, Dong, et al. [11] used Pyramid Vision Transformers (PVT) as an encoder to extract more comprehensive feature information. J. Chen et al. [12] proposed a U-shaped codec network TransUNet. TransUNet uses CNN as a feature extractor, and then inputs the extracted feature maps into the Transformer. More deep learning segmentation methods based on neural networks are reviewed in Section II.

On the whole, there are two weaknesses of traditional colonic polyp segmentation methods: 1) These methods are mainly realized based on manually extracted low-level features. Therefore, the segmentation performance is greatly influenced by the expert level. 2) When the polyp size is different, the color contrast between polyps and background is low, especially for small polyps, the overall segmentation performance of these methods is weak. Compared with traditional methods, CNN based colonic polyp segmentation methods can automatically learn features, which reduce the manual error and improve segmentation performance to a certain extent. However, this kind of methods still has the following limitations: 1) They can only establish the short-range dependency, and cannot establish the relationship between target pixel and global pixel. 2) When segmenting small-target colonic polyps, this kind of methods needs to be further improved. Compared with CNN based segmentation methods, Transformer based colonic polyp segmentation methods can establish the association between each pixel. However, this kind of methods misses the ability to capture local features.

In view of the above challenges, a novel hybrid network Fusion-Transformer-HardNetMSEG (Fu-TransHNet) is proposed in this paper. Fu-TransHNet consists of two deep learning branches (i.e., Transformer branch and CNN branch) and a fusion module.In the Transformer branch, the DeiT-small [13] is introduced to realize global feature learning of colonic polyp images. In the CNN branch, the lightweight network HardNet-MSEG [14] is adopted to achieve local feature learning of colonic polyp images. 3) In the fusion module, the global and local features obtained from the two branches are fused.The contributions of this paper are summarized as follows:

(1) A hybrid network Fu-TransHNet is proposed to improve the segmentation accuracy of colonic polyps, especially for small-target colonic polyps. Fu-TransHNet combines the local feature learning ability of CNN and the global feature learning ability of Transformer.

(2) A novel fusion module is designed. Firstly, the global-local feature fusion GLFF in the fusion module, which fuses multi-scale feature maps from CNN and Transformer branches to reduce information loss in position and detail. Secondly, the dense fusion of multi-scale features DFM in the fusion module. DFM densely integrates the fused feature maps of different scales to combine the high-level and low-level features, so as to enhance the information representation.

(3) The weight of each view is obtained adaptively by introducing multi-view cooperation learning to realize the comprehensive decision of these three views.

(4) Numerous experimental analyses, including performance analysis, visualization analysis, comparison analysis of each branch and fusion module, and effectiveness analysis of GLFF and DFM, are organized to verify the effectiveness of the proposed Fu-TransHNet.

The remainder of this paper is organized as follows. Section II reviews classical deep learning methods used for colonic polyp segmentation. The details of the proposed hybrid network Fu-TransHNet is introduced in Section III. Section IV conducts extensive experiments to verify the superior of the proposed Fu-TransHNet. The conclusion of this work is represented in Section V.

## II Related Work

For colonic polyp segmentation, the current mainstream methods are based on deep learning. In this section, we give a review of relevant work.

### A. Colonic polyp segmentation based on CNN learning mechanism

Deep learning methods based on CNN is outstanding in colonic polyp segmentation. Compared with traditional methods, UNet++ [15] and ResUNet++ [16] improved from the classical UNet [17] network have better performance in colonic polyp segmentation. J M et al. [18] applied a conditional adversarial network for colonic polyp segmentation. In this network, the UNet convolutional network is first utilized as a generator to perform the segmentation task, and then the generated segmentation prediction map and the ground-truth are input into the discriminator to identify the true and false. Through the mutual game, the discrimination ability and segmentation accuracy are improved. D. Jha et al. [19] proposed an extended ResUNet++ network structure, which improves the segmentation performance by adding Conditional Random Fields (CRF) and Test Time Augmentation (TTA). In order to solve the boundary ambiguity caused by the similar color between polyps and background, D. P. Fan et al. [20] proposed a convolutional network architecture that uses the reverse attention mechanism to obtain boundary cues. In order to cope with the problem of different sizes and shapes of colonic polyps, P. Song et al. [21] proposed a convolutional network, which introduces the parallel attention modules to achieve multi-scale colonic polyp segmentation.

The CNN based method designed for a certain characteristic of colonic polyps improves the segmentation accuracy to some extent. Since the convolution operation extracts features from adjacent pixels, these methods only realize the short-range dependency, but ignore the long-range dependency. Therefore, these CNN-based colonic polyp segmentation methods face a bottleneck in improving the segmentation accuracy.

### B. Colonic polyp segmentation based on Transformer learning mechanism

In order to break the inherent limitations of CNN structure, Transformer network has been introduced into the colonic polyp segmentation in recent years. By directly using self-attention mechanism instead of convolution operation, this kind of methods obtains strong global feature representation.

B, Dong et al. proposed a pyramid vision transformer-based colonic polyp segmentation method [11]. This method integrates the semantic and local information of colonic polyps and obtains representative features.

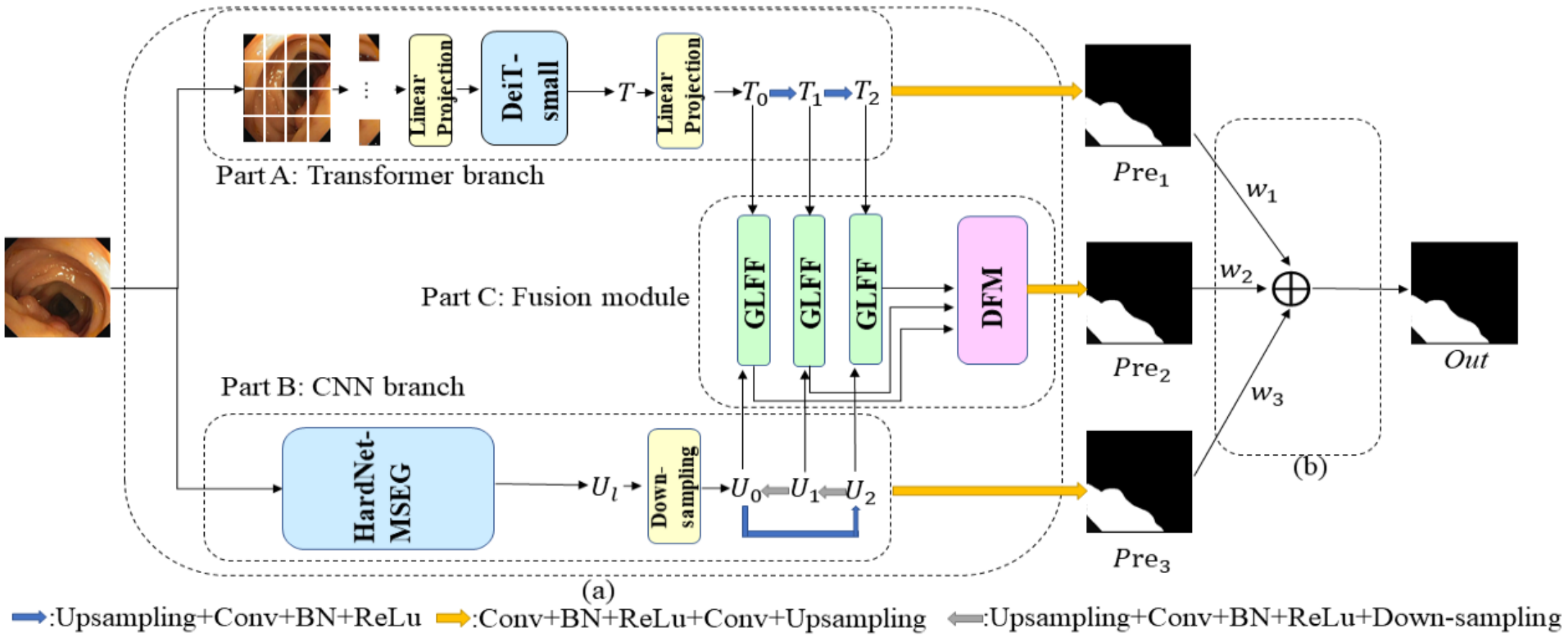

Fig. 1. The pipeline of the Fu-TransHNet: (a) the structure of the Fu-TransHNet; (b) the comprehensive decision of Fu-TransHNet

Compared with CNN-based methods, although the receptive feature learning. Therefore, the colonic polyp segmentation method based on Transformer learning mechanism still needs to be studied.

### C. Colonic polyp segmentation based on Transformer and CNN learning mechanism

Sections II-A and II-B show that colonic polyp segmentation based on CNN learning mechanism and Transformer learning mechanism have their specific advantages and weaknesses. In fact, the two methods can learn from each other. In view of this, some work began to focus on colonic polyp segmentation methods combining Transformer and CNN learning mechanisms. For example, J. Chen et al. proposed a hybrid network by cascading CNN and Transformer (TransUNet) [12]. Specifically, this network uses Transformer to extract global information while preserving the local feature information extracted by CNN. Then the decoder is used to upsample the feature map. Finally, the feature maps are combined by skip connections to achieve accurate positioning. Y. Zhang et al. [22] proposed TransFuse network for colonic polyp segmentation, which combines Vision Transformer and CNN in parallel. By fusing feature maps extracted from the two deep learning mechanisms, TransFuse integrates the local information and the global semantic information.

To improve the performance of colonic polyp segmentation, an effective way is to integrate global perception and local perception by combining CNN and Transformer networks. However, the existing work still faces challenges. For example, TransUNet network first inputs the image into CNN to extract the feature map, which is input into Transformer. This way makes the information obtained by Transformer incomplete and seriously affects the extraction of global information. The output of TransFuse takes into account of the influence of one Transformer and two fusion parts, but ignores the CNN. In addition, TransFuse determines the weights of three outputs existing Transformer-based methods can capture global features of colonic polyps, they do not show obvious advantages in overall segmentation performance. This is mainly due to the shortcomings of these methods in local by manually designing hyperparameters, which makes the network less adaptive.

Although the above methods improve the segmentation accuracy in some datasets, they are still not satisfactory for small-target colonic polyp segmentation. Therefore, we will propose a novel colonic polyp segmentation method based on the efficient Transformer-CNN fusion, to effectively overcome the shortcomings of the existing methods.

## III FU-TRANSHNET

The detail of the proposed Fu-TransHNet is as follows:

### A. Model framework

The framework of the proposed Fu-TransHNet is shown in Fig. 1(a). The core of Fu-TransHNet consists of three parts, that is, the Transformer branch, the CNN branch, and the fusion module. Details of these three parts are as follows:

1) Transformer branch (Part A): This part mainly utilizes the long-range learning ability of Transformer to capture the global information in colonic polyp image. Specifically, the image is first converted into a series of token sequences and then input into DeiT-small for feature extraction. For the output of DeiT-small, the dimension transformation is performed. Further, feature maps corresponding to different scales are obtained by post-processing operations such as the upsampling.

2) CNN branch (Part B): This part mainly adopts the short-range perceptual learning ability of CNN to capture the local information in colonic polyp images. Specifically, we employ the HardNet-MSEG, a lightweight segmentation network using HardNet68 [23], as the backbone for feature extraction. For output of HardNet-MSEG, down-sampling and other post-processing operations are introduced to obtain feature maps corresponding to different scales.

3) Fusion module (Part C): This module includes two parts, namely GLFF and DFM. In order to mitigate the

information loss of feature maps obtained from two branches, GLFF fuses feature maps from two branches under the same scale. Further, in order to fully mine the semantic information under different scales and enhance the feature representation, DFM densely fuses the feature maps from GLFFs under different scales.

Finally, the weights of two branches and one fusion module are learned through multi-view learning to achieve enhanced comprehensive decision.

## B. Global feature learning based on Transformer branch

The Transformer branch of the proposed Fu-TransHNet is shown in Fig. 2(a). The specific learning process of this branch is as follows:

### 1) Pre-processing

In order to perform long-range global feature learning in Transformer branch, colonic polyp images need to be transformed into sequences. The specific flowchart is shown in Fig. 2(a). Firstly, the input image $\mathbf{Y} \in R^{H*W*C}$ is divided into $N = \frac{H}{16} \times \frac{W}{16}$ patches, and then the $N$ patches are flattened into a sequence. Secondly, the sequence is mapped to $d_{model}$ (the dimension of the linear sequence accepted by DeiT-small) by linear transformation. Finally, the position information is added to the sequence and input into DeiT-small.

### 2) DeiT-small

For the colonic polyp segmentation in this paper, the size of the available datasets is not large. In order to achieve superior performance, it is crucial for the proposed Transformer branch to select the appropriate backbone structure. Different from Vision Transformer (ViT) [24], which needs to be pre-trained on large datasets to achieve superior performance, DeiT-small, a Transformer network based on ViT, can obtain excellent performance on small datasets with small-size parameters. Therefore, it is reasonable to adopt the DeiT-small as the core of the proposed Transformer branch. The structure of DeiT-small is shown in Fig. 2(b).

The key of DeiT-small is Multi-head Self-Attention (MSA) and Multi-Layer Perceptron (MLP). MSA maps the input into a rich feature space for parallel processing to extract global features. MLP is mainly used to improve the nonlinear transformation ability of the network. The details of MSA and MLP are described as follows:

#### 1) MSA

The self-attention layer can establish the connection between each token in the sequence. Fu-TransHNet utilizes MSA to focus on multiple key regions simultaneously, instead of the single-head self-attention layer limited to a single feature subspace. In this way, the Fu-TransHNet can obtain richer global feature information of colonic polyps. The small-target colonic polyp segmentation is also proud of this. The process can be expressed as Eqs. (1)-(3).

$$\boldsymbol{Q}_i = \boldsymbol{X}\boldsymbol{W}^{Q_i},\ \boldsymbol{K}_i = \boldsymbol{X}\boldsymbol{W}^{K_i},\ \boldsymbol{V}_i = \boldsymbol{X}\boldsymbol{W}^{V_i} \quad (1)$$

$$\boldsymbol{Head}_i = Attention(\boldsymbol{Q}_i, \boldsymbol{K}_i, \boldsymbol{V}_i), i = [1,2,3 \ldots, h] \quad (2)$$

$$MultiHead(\boldsymbol{Q}, \boldsymbol{K}, \boldsymbol{V}) = Concat(\boldsymbol{Head}_1, \ldots, \boldsymbol{Head}_h)\boldsymbol{W}^{\boldsymbol{o}} \quad (3)$$

where $\boldsymbol{X}$ represents the input matrix. $\boldsymbol{W}^{Q_i}$, $\boldsymbol{W}^{K_i}$ and $\boldsymbol{W}^{V_i}$ are three different weight matrices to be trained in the $i$-th self-attention layer. $h$ represents the number of attention layers. $\boldsymbol{Head}_i$ represents the output matrix of the $i$-th attention layer. $\boldsymbol{W}^{\boldsymbol{o}}$ is the weight matrix of the concating layer.

#### 2) MLP

MLP contains two fully connected layers and a nonlinear layer GELU, whose main function is to perform feature transformation on the input sequence. It is worth noting that the Transformer model uses a residual structure to avoid the vanishing gradient problem. Eq. (4) shows the specific calculation.

$$\boldsymbol{X}' = \boldsymbol{X} + MultiHead(\boldsymbol{Q}, \boldsymbol{K}, \boldsymbol{V}) \quad (4)$$

$$\boldsymbol{T} = MLP(\boldsymbol{X}') + \boldsymbol{X}' \quad (5)$$

where the normalized matrix $\boldsymbol{X}'$ is passed into the MLP to obtain the final output $\boldsymbol{T}$.

As a feature extractor in the Transformer branch, DeiT-small extracts global features of colonic polyps by exploiting the MSA mechanism to establish long-range dependency between pixels. Different from the convolution in the CNN mechanism, the weights in the MSA layer are dynamic and contain the global receptive field. Therefore, the feature map with global information can be obtained. For colonic polyp segmentation, more comprehensive feature information can be obtained in this way, thereby improving the segmentation accuracy.

### 3) Post-processing

In order to integrate the feature maps of the same scale in the Transformer branch and the CNN branch, post-processing operations are performed on the output of DeiT-samll. The post-processing is as follows:

Firstly, for the output sequence $\boldsymbol{T}$ of DeiT-small, it is reshaped into a 2D feature image $\boldsymbol{T}_0(\frac{H}{16} \times \frac{W}{16} \times d_{model})$ in the $d_{model}$ channel. Secondly, the progressive upsampling (i.e., alternating convolution and upsampling) is performed twice to obtain the multi-scale feature maps $\boldsymbol{T}_1(\frac{H}{8} \times \frac{W}{8} \times 128)$ and $\boldsymbol{T}_2(\frac{H}{4} \times \frac{W}{4} \times 64)$.

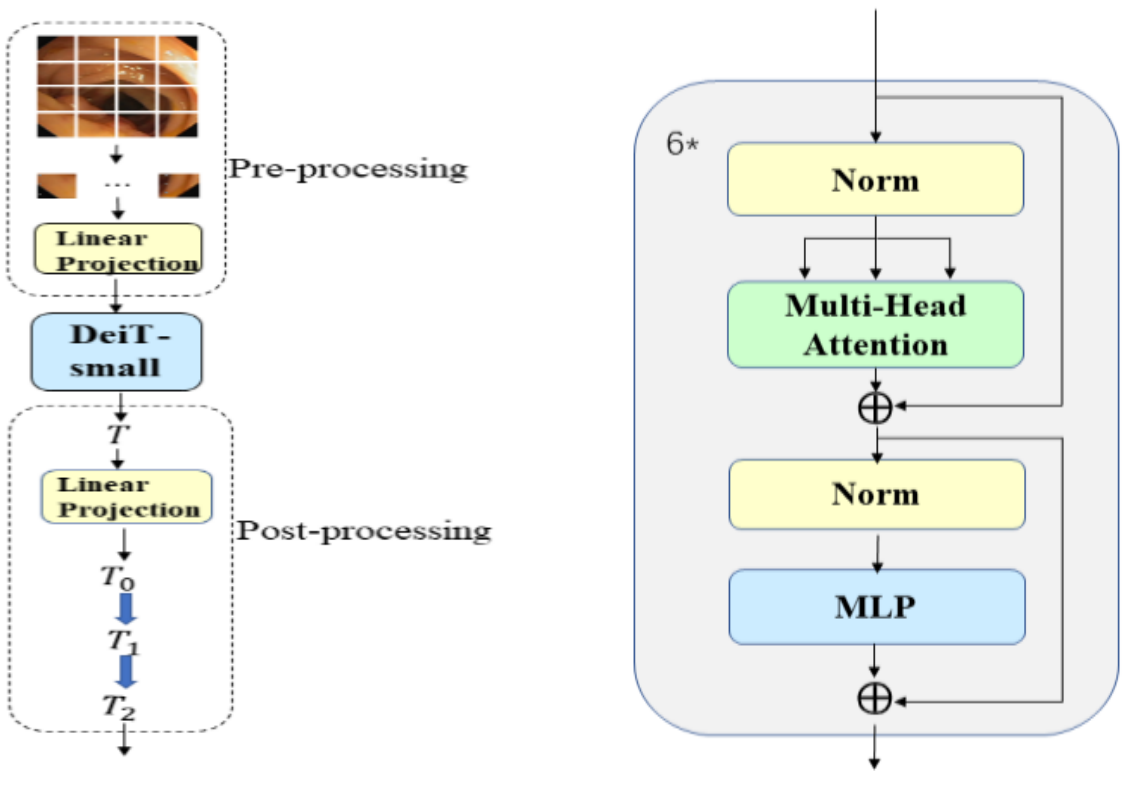

(a) (b)

Fig. 2. The Transformer branch: (a) the flowchart of the Transformer branch; (b) the structure of the DeiT-small.

## C. Local feature learning based on CNN branch

Fig. 3(a) shows the CNN branch of the proposed Fu-TransHNet, where HardNet-MSEG is the core of CNN branch. and its structure is shown in Fig 3(b). The specific learning content is as follows:

### 1) HardNet-MSEG

Due to the different sizes and shapes of colon polyp images, the segmentation accuracy is low. Although some researches [25], [26], [27], [28], [29], [30], [31], [32] have made corresponding progress, the effect is not significant. In this paper, we utilize the local feature learning based on CNN branch to improve the above problem.

On the whole, We chose to use HardNet-MSEG to extract local features for the following four reasons: 1) The HardNet68 network is selected as the backbone of HardNet-MSEG. It mainly uses HarDNet component, i.e., HarDBlk, to speed up the network inference through sparse connections and weighting. This approach allows the network to achieve excellent segmentation performance with little resource consumption. 2) The convolution operation is included in HardNet-MSEG network. It can establish short-range dependency between neighboring pixels to achieve local receptive field. 3) The Receptive Field Block (RFB) [33] component is added to the HardNet-MSEG. By increasing the receptive field, it can obtain more feature information. In addition, it can also improve the limitation caused by the lack of global receptive field in CNN structure. 4) The Aggregation module is introduced to aggregate the local feature information effectively.

The feature map containing more refined features is helpful to improve the segmentation performance of Fu-TransHNet.

### 2) Post-processing

In order to effectively fuse the local feature information of CNN branch with the global feature information of Transformer branch, it is necessary to post-process the output of HardNet-MSEG. Firstly, the output of HardNet-MSEG $\boldsymbol{U}_l$ is down-sampled to get $\boldsymbol{U}_0$. Secondly, $\boldsymbol{U}_2(\frac{H}{4}\times\frac{W}{4}\times 64)$ is obtained by double upsampling, convolution, BatchNorm and ReLU on $\boldsymbol{U}_0$. Thirdly, $\boldsymbol{U}_1(\frac{H}{8}\times\frac{W}{8}\times 128)$ is obtained by upsampling and down-sampling on $\boldsymbol{U}_2(\frac{H}{4}\times\frac{W}{4}\times 64)$. Similarly, $\boldsymbol{U}_1$ is updated to $\boldsymbol{U}_0(\frac{H}{16}\times\frac{W}{16}\times 256)$.

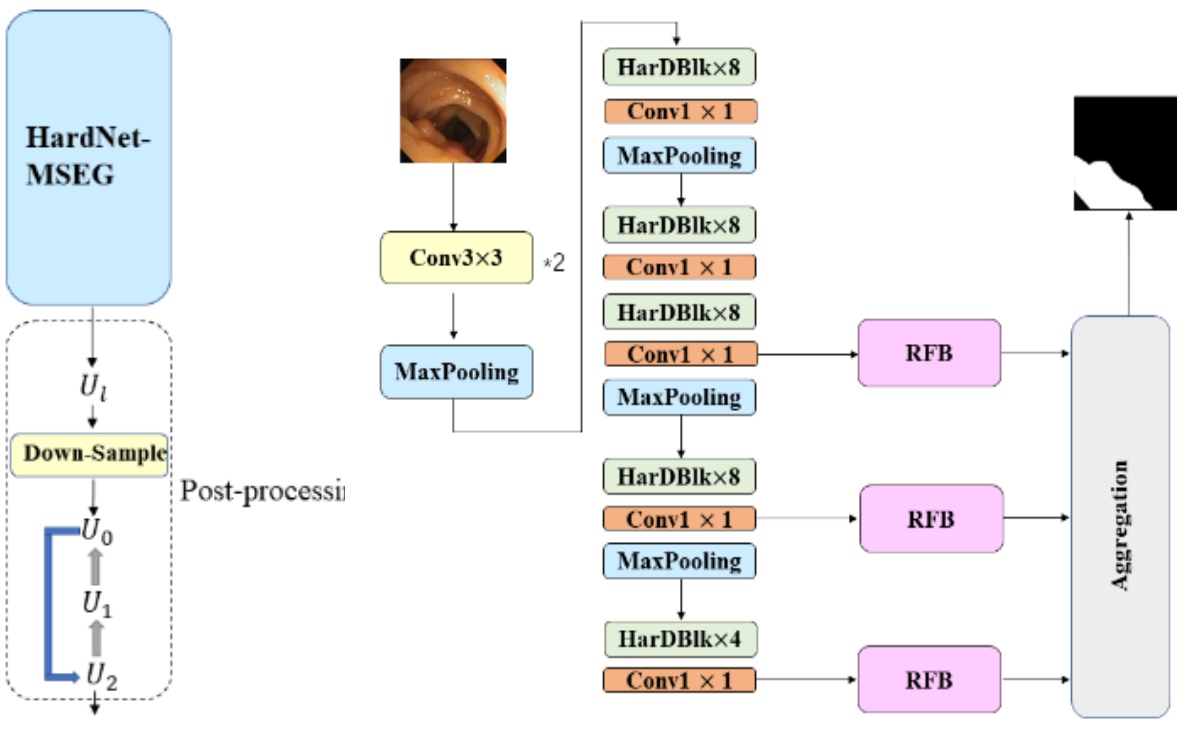

Fig. 3. The CNN branch: (a) the flowchart of the CNN branch; (b) the structure of the HardNet-MSEG.

## D. Fusion module

In order to combine the advantages of the Transformer branch and CNN branch, we design a fusion module in the proposed hybrid network Fu-TransHNet. The fusion module consists of two parts: GLFF and DFM. The former generates feature maps by fusing the global and local features under the same scale of two branches. The latter densely integrates the multi-scale feature maps obtained by the former. The combination of GLFF and DFM reduces the information loss and improves the segmentation accuracy of small-target colonic polyps.

### 1) Global-local feature fusion (GLFF)

In this work, we used three different scales for global-local feature fusion, as shown in Part C of Fig. 1. In addition, Fig. 4 gives the specific structure of GLFF. Specifically, the feature maps obtained from the Transformer and the CNN branches under the same scale are fused by GLFF. GLFF consists of three steps:

Firstly, the Convolutional Block Attention Module (CBAM) [34] is used to focus on the channel and spatial features of the feature maps (i.e., $\boldsymbol{T}_i$ and $\boldsymbol{U}_i$) from the two branches. In this way, background information of colonic polyps is suppressed, and redundant information is reduced.

Secondly, the *Concat* channel connection is performed on the feature maps of the two branches. And then the convolution operation is performed to extract the combined features from the multi-channel input.

Thirdly, the Softmax activation function is applied to the feature map to obtain a pixel-level probability map $\boldsymbol{\alpha}_i$ and $\boldsymbol{\beta}_i$, which is respectively multiplied with the feature maps from the two branches. The obtained results are then summed, and the final fusion feature map $\boldsymbol{F}_i$ is obtained using the residual connection.

The above process is formulated as Eqs. (6)-(7).

$$\boldsymbol{F}_i = Residual[CBAM(\boldsymbol{T}_i)\odot\boldsymbol{\alpha}_i + CBAM(\boldsymbol{U}_i)\odot\boldsymbol{\beta}_i],\ i=[0,1,2] \tag{6}$$

$$\boldsymbol{\alpha}_{i,j}=\frac{e^{T_{i,j}}}{e^{U_{i,j}}+e^{T_{i,j}}},\ \boldsymbol{\beta}_{i,j}=\frac{e^{U_{i,j}}}{e^{U_{i,j}}+e^{T_{i,j}}},\ j=[1,2,3\ldots,H\times W] \tag{7}$$

where $\odot$ is the element-wise product. $\boldsymbol{\alpha}_i$, $\boldsymbol{\beta}_i$ are the probability map of $\boldsymbol{T}_i$ and $\boldsymbol{U}_i$, respectively. $\boldsymbol{\alpha}_{i,j}$, $\boldsymbol{\beta}_{i,j}$ represents the probability value at the *j*-th pixel in the *i*-th probability map.

GLFF fuses the information obtained from the two branches to avoid discarding some edge information that is crucial for colon polyp (especially small-target colonic polyp) segmentation.

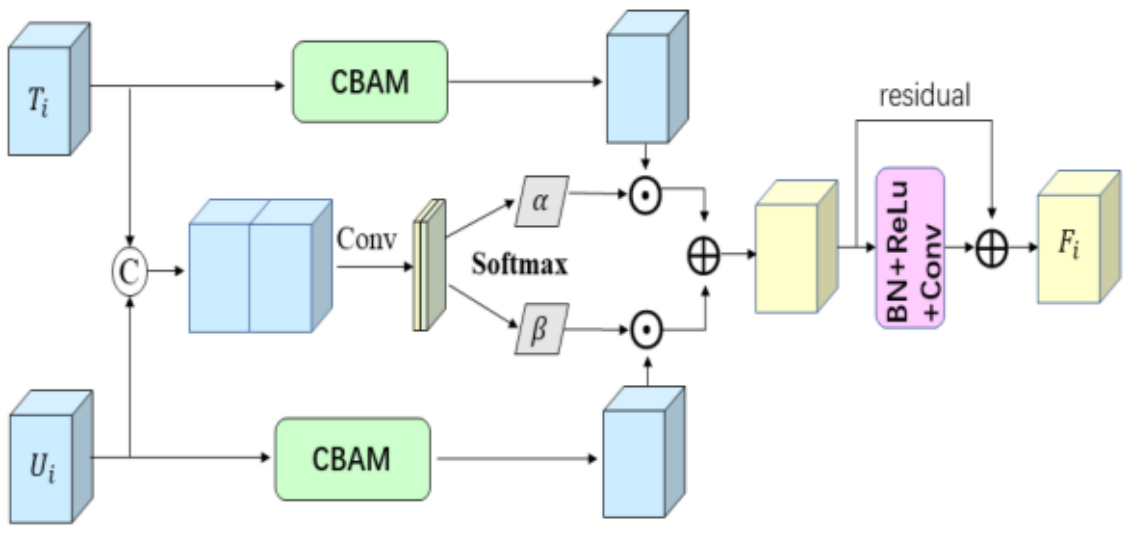

Fig. 4. The structure of the GLFF ($\odot$ is element-wise product and $\oplus$ is element-wise add).

### 2) Dense fusion of multi-scale features (DFM)

According to the Section III-D-1, we will get feature maps under different scales. Among them, high-level feature map has rich semantic information, and low-level feature map contains much position and edge information due to the high resolution. In order to obtain the comprehensive feature representation, the DFM is introduced to perform dense fusion on the output $\boldsymbol{F}_i$ of GLFFs. The structure of DFM is shown in Fig 5.

In order to effectively detect small-target colonic polyps, DFM densely fuses the high-level, mid-level and low-level feature maps $\boldsymbol{F}_0, \boldsymbol{F}_1, \boldsymbol{F}_2$ at different scales. In this way, Fu-TransHNet can effectively avoid missing information of small-target colonic polyps. The specific process of DFM is as follows:

Firstly, the feature map $\boldsymbol{F}_0$ is executed with the UniT operation to obtain $\boldsymbol{F}_{0-1}$. Secondly, $\boldsymbol{F}_{1-1}$ can be gotten by computing the element-by-element summation between $\boldsymbol{F}_{0-1}$ and mid-level feature map $\boldsymbol{F}_1$, which is calculated by convolution. Thirdly, $\boldsymbol{F}_{0-1}$ is *Concat* with $\boldsymbol{F}_{1-1}$ operated by UniT to obtain $\boldsymbol{F}_{1-2}$. Similar operations are performed on $\boldsymbol{F}_1$ and $\boldsymbol{F}_2$. Finally, the output $\boldsymbol{Out}$ is obtained by performing UniT twice on the feature map $\boldsymbol{F}_{2-2}$.

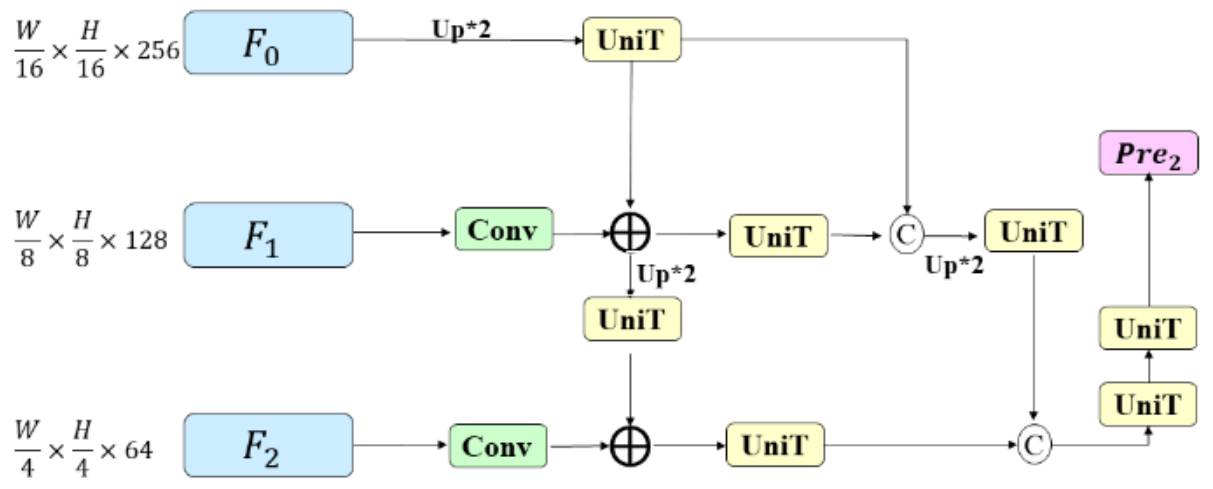

Fig. 5. The structure of the DFM (UniT is represented as a convolution unit consisting of 3 × 3 convolution, BatchNorm, and ReLU. Up × 2 is represented as a simple 2-fold upsampling operation, © is a *concat* operation, conv is a convolution operation, and ⊕ is element-wise add).

### *E. Multi-view cooperation based objective function and its optimization*

In the proposed Fu-TransHNet, the Transformer branch, CNN branch and the fusion module will generate three different predicted segmentation results. To obtain the effective segmentation performance, these predicted results should be considered comprehensively. The simplest strategy is to treat each of them equally, or assign weights to them manually [20][22]. However, these strategies are crude and less adaptive. A reasonable approach is to adaptively evaluate their importance during the learning process.

Based on the above analysis, we introduce the multi-view cooperation enhancement to train the proposed Fu-TransHNet. Therefore, the three prediction results of Transformer branch, CNN branch and fusion module are regarded as three views. That is, view 1 (Transformer branch), view 2 (CNN branch), and view 3 (fusion module). The details of the objective function based on multi-view cooperation are as follows:

Firstly, the multi-view cooperation based objective function is

$$\min Loss(\boldsymbol{w}, \boldsymbol{\Theta}) = \sum_{k=1}^{K} w_k\, loss_k + \lambda \sum_{k=1}^{K} w_k \ln w_k$$
$$s.t.\ \sum_{k=1}^{K} w_k = 1 \qquad (8)$$
$$loss_k = loss(\boldsymbol{Pre}_k, \boldsymbol{G}) \qquad (9)$$

where $\boldsymbol{w} = [w_1, \dots, w_v, \dots, w_K]$ is a weight vector corresponding to $K$ views. $w_k$ is the $k$-th view weight. $K$ is the total number of views. In this paper, $K$=3. $\boldsymbol{\Theta}$ is the set of model parameters to be learned. $Loss$ is the total objective function. $loss_k$ is the loss of the $k$-th view. $\boldsymbol{Pre}_k$ is the prediction map of the $k$-th view. $\boldsymbol{G}$ is the ground-truth. $\lambda$ is a hyperparameter.

For the Eq. (8), it adopts the weighting mechanism based on Shannon entropy, which can effectively adjust the importance of each view. And it can be solved by alternating iteration. The specific steps are as follows:

1) When $\boldsymbol{\Theta}$ is fixed, $w_k$ can be solved by the Lagrangian optimization. The optimal solution of $w_k$ can be expressed as:

$$w_k = e^{\frac{-loss_k}{\lambda}} / \sum_{h=1}^{K} e^{\frac{-loss_h}{\lambda}} \qquad (10)$$

2) When $w_k$ is fixed, $\boldsymbol{\Theta}$ in Eq. (8) is updated using the Adam optimizer.

Further, the weighted IoU loss $\mathcal{F}_{IoU}^{w}$ and the binary cross-entropy loss $\mathcal{F}_{bce}^{w}$ participate in the loss measure of Eq. (11).

$$loss(\boldsymbol{Pre}_k, \boldsymbol{G}) = \mathcal{F}_{IoU}^{w} + \mathcal{F}_{bce}^{w} \qquad (11)$$

The specific calculation of $\mathcal{F}_{IoU}^{w}$ and $\mathcal{F}_{bce}^{w}$ is shown in Eqs. (12) and (13):

$$\mathcal{F}_{IoU}^{w} = 1 - \sum_n \frac{y_n \times p_n}{y_n + p_n - y_n \times p_n} \qquad (12)$$

$$\mathcal{F}_{bce}^{w} = -\sum_n (y_n \times \log(p_n) + (1 - y_n) \times \log(1 - p_n)) \qquad (13)$$

where $y_n$ represents the true label of the $n$-th pixel value, with positive sample labeled as 1 and negative sample labeled as 0. $p_n$ is the probability that the $n$-th pixel value is predicted to be a positive sample.

For the proposed objective function based on multi-view cooperation enhancement, it can adaptively obtain the view weights without manual adjustment. Therefore, the proposed network Fu-TransHNet has good adaptability.

Based on the above analysis, Table I summarizes the algorithm description of the proposed Fu-TransHNet.

Table I Algorithm description of Fu-TransHNet

| **Algorithm:** Fu-TransHNet | |
|---|---|
| **Input:** | Training images with labels $\boldsymbol{Y}$, the maximum iteration number $T$ =200. |
| **Output:** | Segmentation prediction map |
| **Procedure:** | |
| 1: | Initialize model parameters, $t$=0 |
| 2: | **While** *not converge* or $t<T$ |
| 3: | Fix $\boldsymbol{\Theta}$, using Eqs. (10)-(13) to solve $w_k$. |
| 4: | Fix $w_k$, using the Adam optimizer to update the model parameters $\boldsymbol{\Theta}$. |
| 5: | $t$ = $t$+1. |
| 6: | **End while** |

### *F. Comprehensive decision of Fu-TransHNet*

Most multi-branch networks select only one branch as the basis for final prediction, which is crude, such as [19][35]. Different from this, we propose the comprehensive decision to obtain the optimal prediction, as shown in Fig. 1(b).

First, the weight of each view is calculated by multi-view weighting in the proposed Fu-TransHNet. The final predicted output is then computed by the weighted summation of all views. Eq. (14) gives the specific formula:

$$\boldsymbol{Out} = \sum_{k=1}^{K} w_k \times \boldsymbol{Pre}_k \qquad (14)$$

For colonic polyp segmentation, it is expected to achieve superior performance by cooperating with the prediction of three views (i.e., the Transformer branch, the CNN branch and the fusion module).

Table II Performance of 11 methods on CVC-ClinicDB dataset

| Method | Evaluation metric | | | | | |
|---|---|---|---|---|---|---|
| | mDice | mIou | $F_{\beta}^{w}$ | $S_{\alpha}$ | $E_{\phi}^{max}$ | MAE |
| UNet | 0.823 | 0.755 | 0.811 | 0.889 | 0.954 | 0.019 |
| UNet++ | 0.794 | 0.729 | 0.785 | 0.873 | 0.931 | 0.022 |
| SFA | 0.700 | 0.607 | 0.647 | 0.793 | 0.885 | 0.042 |
| PraNet | 0.899 | 0.849 | 0.896 | 0.936 | 0.979 | 0.009 |
| HardNet-MSEG | 0.901 | 0.850 | 0.892 | 0.929 | 0.971 | 0.009 |
| SANet | 0.916 | 0.859 | 0.909 | 0.939 | 0.976 | 0.012 |
| AMNet | **0.936** | **0.888** | - | - | - | **0.007** |
| Polyp-pvt | 0.919 | 0.871 | 0.916 | 0.938 | 0.974 | 0.011 |
| TransFuse-s | 0.896 | 0.845 | 0.890 | 0.921 | 0.958 | 0.014 |
| TransUNet | 0.935 | 0.887 | - | - | - | - |
| Fu-TransHNet | 0.926 | 0.876 | **0.921** | **0.944** | **0.980** | 0.008 |

Table III Performance of 11 methods on CVC-ColonDB dataset

| Method | Evaluation metric | | | | | |
|---|---|---|---|---|---|---|
| | mDice | mIou | $F_{\beta}^{w}$ | $S_{\alpha}$ | $E_{\phi}^{max}$ | MAE |
| UNet | 0.512 | 0.444 | 0.498 | 0.712 | 0.776 | 0.061 |
| UNet++ | 0.483 | 0.410 | 0.467 | 0.691 | 0.760 | 0.064 |
| SFA | 0.469 | 0.347 | 0.379 | 0.634 | 0.765 | 0.094 |
| PraNet | 0.709 | 0.640 | 0.696 | 0.819 | 0.869 | 0.045 |
| HardNet-MSEG | 0.732 | 0.656 | 0.709 | 0.824 | 0.877 | 0.039 |
| SANet | 0.753 | 0.670 | 0.726 | 0.837 | 0.878 | 0.043 |
| AMNet | 0.762 | 0.690 | - | - | - | 0.033 |
| Polyp-pvt | 0.778 | 0.701 | 0.757 | 0.845 | 0.886 | 0.046 |
| TransFuse-s | 0.769 | 0.692 | 0.751 | 0.845 | 0.876 | 0.035 |
| TransUNet | 0.781 | 0.699 | - | - | - | - |
| Fu-TransHNet | **0.810** | **0.733** | **0.800** | **0.866** | **0.909** | **0.030** |

Table IV Performance of 11 methods on CVC-EndoScene dataset

| Method | Evaluation metric | | | | | |
|---|---|---|---|---|---|---|
| | mDice | mIou | $F_{\beta}^{w}$ | $S_{\alpha}$ | $E_{\phi}^{max}$ | MAE |
| UNet | 0.710 | 0.627 | 0.684 | 0.843 | 0.875 | 0.022 |
| UNet++ | 0.707 | 0.624 | 0.687 | 0.839 | 0.898 | 0.018 |
| SFA | 0.467 | 0.329 | 0.341 | 0.640 | 0.817 | 0.065 |
| PraNet | 0.871 | 0.797 | 0.843 | 0.925 | 0.972 | 0.010 |
| HardNet-MSEG | 0.878 | 0.810 | 0.854 | 0.925 | 0.957 | 0.009 |
| SANet | 0.888 | 0.815 | 0.859 | 0.928 | 0.972 | 0.008 |
| Polyp-pvt | 0.898 | 0.830 | 0.879 | 0.934 | 0.974 | 0.008 |
| TransFuse-s | 0.870 | 0.797 | 0.844 | 0.916 | 0.943 | 0.010 |
| TransUNet | 0.893 | 0.824 | - | - | - | - |
| Fu-TransHNet | **0.903** | **0.834** | **0.883** | **0.933** | **0.969** | **0.006** |

Table V Performance of 11 methods on ETIS- LaribpolypDB dataset

| Method | Evaluation metric | | | | | |
|---|---|---|---|---|---|---|
| | mDice | mIoU | $F_{\beta}^{w}$ | $S_{\alpha}$ | $E_{\phi}^{max}$ | MAE |
| UNet | 0.398 | 0.335 | 0.366 | 0.684 | 0.740 | 0.036 |
| UNet++ | 0.401 | 0.344 | 0.390 | 0.683 | 0.776 | 0.035 |
| SFA | 0.297 | 0.217 | 0.231 | 0.557 | 0.633 | 0.109 |
| PraNet | 0.628 | 0.567 | 0.600 | 0.794 | 0.841 | 0.031 |
| HardNet-MSEG | 0.666 | 0.592 | 0.632 | 0.809 | 0.863 | 0.019 |
| SANet | 0.750 | 0.654 | 0.685 | 0.849 | 0.897 | 0.015 |
| AMNet | 0.756 | 0.679 | - | - | - | 0.038 |
| Polyp-pvt | 0.760 | 0.680 | 0.719 | 0.852 | 0.892 | 0.025 |
| TransFuse-s | 0.728 | 0.653 | 0.698 | 0.837 | 0.864 | 0.020 |
| TransUNet | 0.731 | 0.660 | - | - | - | - |
| Fu-TransHNet | **0.798** | **0.719** | **0.773** | **0.873** | **0.906** | **0.012** |

Table VI Performance of 11 methods on Kvasir dataset

| Method | Evaluation metric | | | | | |
|---|---|---|---|---|---|---|
| | mDice | mIoU | $F_{\beta}^{w}$ | $S_{\alpha}$ | $E_{\phi}^{max}$ | MAE |
| UNet | 0.818 | 0.746 | 0.794 | 0.858 | 0.893 | 0.055 |
| UNet++ | 0.821 | 0.743 | 0.808 | 0.862 | 0.910 | 0.048 |
| SFA | 0.723 | 0.611 | 0.670 | 0.782 | 0.849 | 0.075 |
| PraNet | 0.898 | 0.840 | 0.885 | 0.915 | 0.948 | 0.030 |
| HardNet-MSEG | 0.905 | 0.847 | 0.886 | 0.912 | 0.951 | 0.029 |
| SANet | 0.904 | 0.847 | 0.892 | 0.915 | 0.953 | 0.028 |
| AMNet | 0.912 | 0.865 | - | - | - | 0.028 |
| Polyp-pvt | 0.912 | 0.859 | 0.901 | 0.918 | 0.955 | 0.026 |
| TransFuse-s | 0.908 | 0.856 | 0.900 | 0.914 | 0.956 | 0.026 |
| TransUNet | 0.913 | 0.857 | - | - | - | - |
| Fu-TransHNet | **0.916** | **0.866** | **0.915** | **0.920** | **0.962** | **0.023** |

## IV EXPERIMENTS

We organize numerous experiments to verify the effectiveness of the proposed Fu-TransHNet, including performance analysis, visualization analysis, comparison analysis of single branch and multi-branch with fusion module, and effectiveness analysis of GLFF and DFM. Datasets and preprocessing, experimental settings, and evaluation metrics are described below.

### A. Datasets and preprocessing

We introduce five published colonic polyp segmentation datasets in this paper, namely CVC-ClinicDB [36], CVC-

ColonDB [37], CVC-EndoScene [38], ETIS-LaribPolypDB [39] and Kvasir [40]. For the sake of fairness, the settings of the training and test sets are the same as [14], [20], and [22]. That is, 900 images in Kvasir and 550 images in CVC-ClinicDB are collected together to construct a training set of 1450 images. The rest of the five datasets (a total of 798 images) are served as the test set. In the experiment, all images are uniformly transformed into 352×352 size.

### B. Experimental settings

The proposed Fu-TransHNet is implemented based on the Pytorch framework. We adopt NVIDIA GTX3090 for GPU acceleration training. The Adam optimizer is utilized to update the model parameters, the learning rate is set to 7e-5, and the epoch is set to 200. The optimal value of hyperparameter $\lambda$ can be obtained by cross-validation or other strategies. In this paper, we set $\lambda = 1$.

For full evaluation, the proposed Fu-TransHNet is compared with 10 existing methods, including 7 colonic polyp segmentation methods based on CNN mechanism (Unet [17], UNet++ [15], SFA [9], PraNet [20], SANet [41], HardNet-MSEG [14] and AMNet [21]), 1 colonic polyp segmentation method based on Transformer (Polyp-pvt [11]), and 2 colonic polyp segmentation methods based on the combination of CNN and Transformer (TransFuse-s [22] and TransUNet [12]). It should be noted that HardNet-MSEG, Polyp-pvt and TransFuse-s are run locally based on the code provided by the authors. The results of other comparison methods are from [11], [22], [35], and [41].

### C. Evaluation metrics

Six evaluation metrics are introduced in this paper, including mean Dice (mDice), mean IoU (mIoU), weighted F-measure ($F_\beta^w$) [42], Structure-measure ($S_\alpha$) [43], max-E-measure ($E_\emptyset^{max}$) [44], and mean absolute error (MAE) [45]. The first two metrics are commonly used in semantic segmentation, and the last four metrics are commonly used in object detection.

### D. Experimental analysis

#### 1) Performance analysis

According to the performance of 11 methods on 5 benchmark datasets shown in Tables II-VI, we can get the following observations:

(1) The comparison of 11 methods on the ETIS-LaribPolypDB dataset (containing many small-target colonic polyps) show that Fu-TransHNet outperforms the state-of-the-art methods HardNet-MSEG and TransFuse-s. Specifically, the mDice of Fu-TransHNet is 13.2% higher than that of HardNet-MSEG, 7.0% higher than that of TransFuse-s, and 3.8% higher than that of the suboptimal Polyp-pvt method. The comparison results demonstrate the superiority of the proposed Fu-TransHNet in small-target colonic polyp segmentation.

(2) According to the results of 11 methods on the CVC-ColonDB, CVC-EndoScene and Kvasir datasets, the proposed Fu-TransHNet has the best performance in all metrics. The mDice of Fu-TransHNet on CVC-ColonDB dataset is 4.8% higher than that of the AMNet network, 3.2% higher than that of the Polyp-pvt network, and 2.9% higher than that of the TransUNet network.

(3) The comparison results of 11 methods on the CVC-ClinicDB dataset show that the mDice of Fu-TransHNet are about 2% better than those of the Polyp-pvt. Compared with AMNet , the proposed Fu-TransHNet is competitive.

On the whole, Fu-TransHNet shows great competitiveness and advantages, especially in small-target colonic polyp segmentation. The superior of the Fu-TransHNet is mainly attributed to the design of a novel fusion module and the introduction of multi-view cooperative learning enhancement.

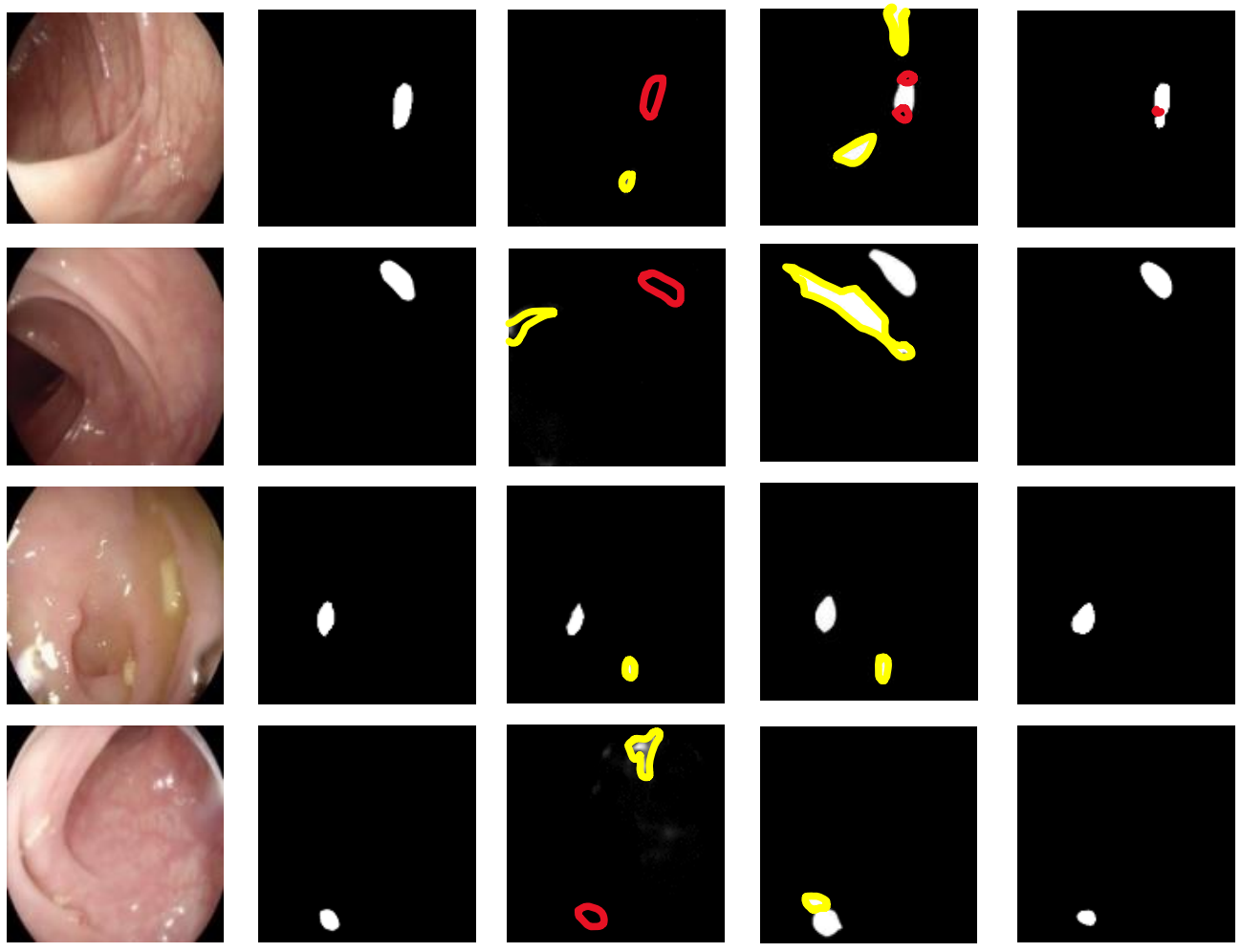

(a) (b) (c) (d) (e)

Fig. 6. Visualization of segmentation results of several methods in ETIS-LaribPolypDB dataset. (a) original. (b) ground truth. (c) the result of HardNet-MSEG, (d) TransFuse-s, (e) Fu-TransHNet.

#### 2) Visualization analysis

In order to verify the advantage of the Fu-TransHNet in small-target colonic polyp segmentation, we visualize the segmentation prediction of four images in the ETIS-LaribPolypDB dataset, which contains many small-target colonic polyps. Since Fu-TransHNet is an improvement on HardNet-MSEG and TransFuse-s, we compare Fu-TransHNet with these two advanced methods. Fig. 6 shows the segmentation prediction of these methods. Specifically, Fig. 6 (a)-(e) show the image, ground-truth, segmentation prediction of HardNet-MSEG, segmentation prediction of TransFuse-s, and segmentation prediction of Fu-TransHNet, respectively. The red mark indicates missed detection, and yellow mark indicates false detection (that is, normal tissue is detected as polyp).

It can be seen from Fig. 6 that for the small-target colonic polyp segmentation, the segmentation predictions of the proposed Fu-TransHNet are basically consistent with the ground-truth. However, HardNet-MSEG is prone to missed detection and false detection, and TransFuse-s is prone to false detection.

The above analysis shows the superior of Fu-TransHNet in small-target colonic polyp segmentation.

Table VII The mDice of Transformer branch, CNN branch, Transformer+CNN, and Fu-TransHNet on five datasets

| Method | Dataset | | | | |
|---|---|---|---|---|---|
| | CVC-ClinicDB | CVC-ColonDB | CVC-EndoScene | ETIS- LaribPolypDB | Kvasir |
| Transformer branch | 0.869 | 0.759 | 0.848 | 0.699 | 0.897 |
| CNN branch | 0.891 | 0.707 | 0.902 | 0.634 | 0.891 |
| Transformer+CNN | 0.907 | 0.788 | 0.885 | 0.745 | 0.902 |
| Fu-TransHNet | **0.926** | **0.810** | **0.903** | **0.798** | **0.916** |

Table VIII Effectiveness(mDice) of GLFF and DFM on five datasets

| Module | | Dataset | | | | |
|---|---|---|---|---|---|---|
| GLFF | DFM | CVC-ClinicDB | CVC-ColonDB | CVC-EndoScene | ETIS- LaribPolypDB | Kvasir |
| - | - | 0.907 | 0.788 | 0.885 | 0.745 | 0.902 |
| √ | - | 0.913 | 0.797 | 0.887 | 0.759 | 0.910 |
| - | √ | 0.912 | 0.795 | 0.892 | 0.772 | 0.914 |
| √ | √ | **0.926** | **0.810** | **0.903** | **0.798** | **0.916** |

### 3) Comparison analysis of each branch and fusion module

Fu-TransHNet is a hybrid network, which utilizes the proposed fusion module to combine the Transformer and CNN branches, so that local features can be extracted through CNN and global information can be obtained through Transformer.

In order to prove the effectiveness of the proposed fusion module, a comparative experiment is organized in this section. Specifically, for Transformer branch, it only uses the Transformer learning mechanism to segment colonic polyps. For CNN branch, it only uses the CNN learning mechanism to segment. For Transformer+CNN, it performs weighted summation of prediction results of Transformer branch and CNN branch. The experimental results are shown in Table VII. By observing Table VII, the following conclusions can be drawn:

(1) The segmentation performance of Transformer+CNN is better than Transformer branch and CNN branch on the whole. This indicates that the simple fusion of the two branches can help to improve the performance of colonic polyp segmentation.

(2) Fu-TransHNet outperforms Transformer+CNN on the five datasets. This shows that the proposed fusion module is beneficial to further improve the segmentation performance.

### 4) Effectiveness analysis of GLFF and DFM

In order to verify the effectiveness of GLFF and DFM, we set up three comparison methods for the proposed Fu-TransHNet in this section. "√" means the corresponding part is valid, and "-" means the corresponding part is invalid. There are three comparison methods, Therefore, there are three comparison methods, namely, Fu-TransHNet_noDFM_noGLFF, Fu-TransHNet_noDFM, and Fu-TransHNet_noGLFF. Table VIII shows the experimental results of the four methods on the mDice. Observing the Table VIII we have the following findings:

(1) The mDice of Fu-TransHNet_noDFM is better than that of Fu-TransHNet_noDFM_noGLFF. This is mainly due to the following two reasons: 1) The CBAM attention mechanism in GLFF passes the input feature map to the channel attention and spatial attention modules. This mechanism greatly improves the feature dependency between space and channel. 2) The convolution in GLFF is used to act on the combined feature map of the two branches, to extract the global-local combined features. This approach reduces the loss of small-target information and thus obtains more abundant feature information.

(2) The mDice of Fu-TransHNet_noGLFF is better than that of Fu-TransHNet_noDFM_noGLFF. This is because DFM combines the low-level feature map and high-level feature map. Low-level feature map has a smaller receptive field, but the representation ability of geometric detail information is strong. High-level feature map has a large receptive field, and the semantic information representation ability is strong. Therefore, DFM can generate stronger feature representation and further improve the segmentation performance of colonic polyps.

(3) The mDice of Fu-TransHNet is better than that of Fu-TransHNet_noDFM_noGLFF, Fu-TransHNet_noDFM, and Fu-TransHNet_noGLFF. This is because the effective combination of GLFF and DFM makes the proposed Fu-TransHNet fully develop the advantages of GLFF and DFM.

The above experimental analysis proves that the GLFF and DFM designed in this paper are reasonable and effective.

## V CONCLUSION

In order to improve the segmentation performance of colonic polyps, especially small-target colonic polyps, a hybrid network Fu-TransHNet is proposed in this paper. Especially, a novel feature fusion module is designed to take full advantage of the local and global features obtained from CNN and Transformer learning mechanisms. For the fusion module, the global-local feature fusion GLFF is constructed firstly to integrate global and local features from Transformer and CNN branches. And then, a dense fusion of multi-scale features DFM is built to further fuse the information from GLFFs under different scales. Finally, the multi-view cooperation enhancement is proposed to adaptively obtain the weights of two branches and a fusion module, which is flexible to achieve optimal performance. Compared with some existing deep learning-based colonic polyp segmentation networks, the proposed network can extract richer feature information and overcome the defect that small-target polyps are easy to be ignored. Extensive experimental analysis verifies the superiority of the proposed Fu-TransHNet network.

For the proposed Fu-TransHNet, there is still some room for further improvement. For example, in the Transformer branch, the image is first segmented into a series of patches and then linearly projected into a series of tokens. This will destroy the original spatial structure of the image, resulting in information loss. In the future, we will conduct in-depth research on the above problem and propose effective improvement strategies. In addition, for the parameter λ in multi-view objective function, its optimal value can be found according to hyperparameter search or other methods to further improve the segmentation performance.